\documentclass[conference]{IEEEtran}
\IEEEoverridecommandlockouts
\usepackage{cite}
\usepackage{amsmath,amssymb,amsfonts}
\usepackage{algorithmic}
\usepackage{graphicx}
\usepackage{textcomp}
\usepackage{xcolor}
\def\BibTeX{{\rm B\kern-.05em{\sc i\kern-.025em b}\kern-.08em
    T\kern-.1667em\lower.7ex\hbox{E}\kern-.125emX}}
\begin{document}

\title{Diabetic Retinopathy Classification using Downscaling Algorithms and Deep Learning
}

\author{\IEEEauthorblockN{Nishi Doshi, Urvi Oza, and Pankaj Kumar}%
\IEEEauthorblockA{Dhirubhai Ambani Institute of Information and Communication Technology\\
Gandhinagar, India---382007\\
\texttt{201601408@daiict.ac.in}, \texttt{201921009@daiict.ac.in}, \texttt{pankaj\_k@daiict.ac.in}}%
}


\maketitle

\begin{abstract}
 Diabetic Retinopathy (DR) is an art and science of recording and classifying the retinal images of a diabetic patient.
DR classification deals with classifying retinal fundus image into five stages on the basis of severity of diabetes. One of the major issue faced while dealing with DR classification problem is the large and varying size of images. In this paper we propose and explore the use of several downscaling algorithms before feeding the image data to a Deep Learning Network for classification. For improving training and testing; we amalgamate two datasets : Kaggle and Indian Diabetic Retinopathy Image Dataset. Our experiments have been performed on a novel Multi Channel Inception V3 architecture with a unique self crafted preprocessing phase.
We report results of proposed approach using accuracy, specificity and sensitivity, which outperform the previous state of the art methods.
\end{abstract}


\section{\textbf{Introduction}}

\begin{figure}[t]
\begin{center}
  \includegraphics[width=0.95\linewidth]{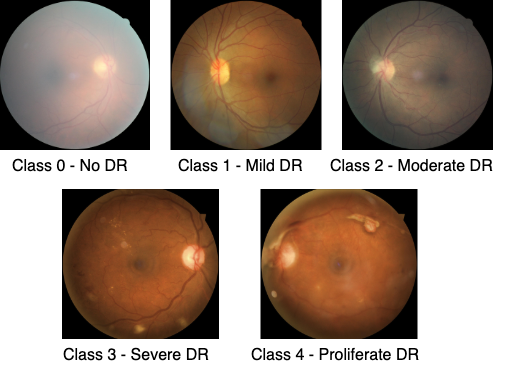}
  \caption{Fundus images belonging to $5$ stages of DR from Kaggle Dataset}
\label{fig4}
\end{center}
\end{figure}

Diabetes is one of the main reasons for blindness in working age of adults\cite{five}. 
The retinal tissue swells, blood veins in retina change which results into their bursting and bleeding. All of this results into blurry vision and loss of eyesight in humans. Early detection of this condition helps in early treatment and prevention of Diabetic Retinopathy (DR). Figure \ref{fig4} shows fundus images belonging to five stages (disease level severity) : no diabetic retinopathy (class $0$), mild (class $1$), moderate (class $2$), severe (class $3$) and proliferate (class $4$).

In order to have bigger dataset and to generalize our results, we propose to amalgamate Kaggle\cite{six} and Indian Diabetic Retinopathy Image Dataset (IDRID)\cite{three} to perform experiments. The dataset used has a large set of high-resolution retina images taken under a variety of imaging conditions. As fundus images are of large and varying sizes; use of downscaling algorithm becomes a major part of prerpocessing. We compare the results of downscaling by different algorithms: nearest neighbour, bilinear, bicubic\cite{twelve}, lanczos \cite{thirteen} and content adaptive down-sampling algorithms: Learned Image Downsampling\cite{one} (LID) and Rapid Detail Image Preserving Downsampling\cite{sixteen} (RDIP). We propose a novel Multi Channel Inception V3 architecture to classify the fundus images and report using metric - accuracy, specificity and sensitivity which are better than existing methods. We bring novelty by experimenting with different downscaling algorithms for preprocessing and proposing a Multi Channel Inception V3 network for classification.

\section{Related Work}
\label{sec1}
Feature Extraction and Deep Learning (DL) models have been used to detect DR in fundus images. Support Vector Machine (SVM) model used to extract spectral features to classify $300$ images into five stages of DR reported sensitivity of $82\%$ and specificity of $88\%$\cite{twenty-one}. SVM classifier was used to classify $400$ images in $4$ classes, after extracting blood vessels from fundus which resulted in $80.4$\% accuracy, $94.6$\% sensitivity and $66.2$\% specificity\cite{twenty-seven}. Algorithms have been proposed to extract blood vessels, exudates, haemorrhages and cloud like structures from retinal images. Classification is done by learning the pattern of these symptoms\cite{twenty-two}. DL methods have been reported to outperform the results of featured based extraction classification.


Convolutional Neural Network (CNN) with $13$ layers was trained for $120$ epochs on $11.654\%$ of Kaggle dataset and for $20$ epochs on entire dataset to avoid over fitting of neural network. It reported $75\%$ accuracy, $95\%$ specificity and $30\%$ sensitivity\cite{seven}.

$9,939$ posterior pole photographs from $2,740$ patients were taken and GoogleNet was applied for classifying fundus images into $14$ classes given by Davis grading system. It reported accuracy of $81$\%. The work also reported false negative rate ($12$\%) and false positive rate ($65$\%) as grade given : not requiring treatment when treatment was actually needed and grade given : requiring treatment but treatment was actually not needed. Thus, mathematically sensitivity and specificity reported were $35$\% and $88$\% respectively\cite{twenty-eight}.

By extracting features and converting RGB images to Gray Scale images; VGG16 and Shallow CNN models were trained separately which resulted in $78.3$\% and $42$\% accuracy respectively\cite{twenty-six}. Binary tree based multi class VGG Network (BT-VGG) was trained on Kaggle dataset which reported $83.2\%$ accuracy, $81.8\%$ sensitivity and $89.3\%$ specificity on $6000$ fundus images\cite{twenty-three}. 

Many research works employed binary classification. Inception V3 network used for binary classification with on $2500$ images from of Kaggle Dataset reported an accuracy of $90.9\%$\cite{twenty}. Pretrained models like Inception V3, Xception, Alexnet, Resnet and VGGNet-s were used for binary classification and among all VggNet-s reported the highest accuracy of $95.68\%$ with hyper-parameter tuning\cite{twenty-four}.
In this paper, we propose a unique preprocessing technique of images and a novel Deep Learning architecture for classifying fundus images into $5$ stages of DR.

\section{Dataset Amalgamation}

We propose to amalgamate two datasets for our experiments - Kaggle Datatset\cite{six} and Indian Diabetic Retinopathy Image Dataset (IDRID)\cite{three}. The datasets have class labels corresponding to $5$ stages of DR - class $0$ label for no DR, class $1$ for mild DR, class $2$ for moderate DR, class $3$ for severe DR and class $4$ for proliferate DR. Both the datasets comprise of images of very large size around $2MB$ of disk space thus making the choice of preprocessing steps of images an important part of classification. A brief description of two datasets is provided in Section \ref{sec2} and Section \ref{sec5}.

\subsection{Kaggle Dataset}
\label{sec2}
A total of $35,216$ images are available from online diabetic retinopathy classification problem on Kaggle\cite{six}. As can be seen in Table \ref{t2}, $73.29\%$ of the images from Kaggle Dataset belong to class $0$ that is no DR. The images obtained from Kaggle have the property of varying sizes as well as different DR severity level for left and right fundus images.

\subsection{Indian Diabetic Retinopathy Image Dataset (IDRID)}
\label{sec5}
A total of $516$ images are available from IDRID\cite{three}. As seen in Table \ref{t2}, $32\%$ images belong to each class $0$ and class $2$ each corresponding to no DR and moderate DR respectively. The images obtained from IDRID are of fixed sizes $4288 \times 2848 \times 3$.

\begin{table}
\begin{center}
\begin{tabular}{|c|c|c|c|}
\hline
Class & Stage & Kaggle Dataset\cite{six} & IDRID Dataset\cite{three}\\
\hline
$0$ & No DR & $25810$ & $168$\\
\hline
$1$ & Mild DR & $2443$ & $25$\\
\hline
$2$ & Moderate DR & $5292$ & $168$\\
\hline
$3$ & Severe DR & $873$ & $93$\\
\hline
$4$ & Proliferate DR & $708$ & $62$\\
\hline
Total & - & $35216$ & $516$\\
\hline
\end{tabular}
\end{center}
\caption{Number of images per class in Kaggle Dataset and IDRID Dataset}
\label{t2}
\end{table}

\section{Proposed Architecture}
DR fundus images are of high resolution, large and varying sizes as well as differ in severity level of left and right eyes. To process such dataset, we propose a two step process of classification. First, we preprocess varying images and transform all images to a fixed size. Then we perform classification on the preprocessed images using DL approach.

\subsection{Preprocessing of Images}

Preprocessing of images include cropping, downscaling and padding before feeding into DL network. The resultant images after preprocessing are of $600 \times 600 \times 3$ size. As we have dataset images of large and varying sizes, down scaling of images becomes a major process. We perform a comprehensive study of downscaling algorithms and evaluate their performance to chose the algorithms which provide the best results. Figure \ref{fig3} shows preprocessing of class $4$ image from Kaggle Dataset.

\subsubsection{Cropping}
\begin{figure}[t]
\begin{center}
  \includegraphics[width=0.95\linewidth]{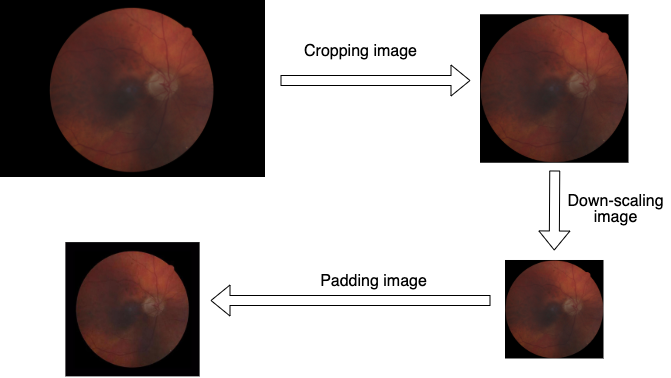}
  \caption{Cropping, Downscaling and Padding preprocessing steps applied on an image}
\label{fig3}
\end{center}
\end{figure}

The images were cropped to concentrate the retina from the available image, as many input images are underexposed or overexposed. As seen from Figure \ref{fig3}, the black portions (in the raw input image from dataset) from all the four sides : top, bottom, right and left are unnecessary parts and cropped off to generate a fundus focused image. The cropped images are of different sizes. The largest size images after cropping are $4800 \times 4800 \times 3$.

\subsubsection{Downscaling}

We downscale the cropped images by $8x$ to feed them to DL network. According to Nyquist-Shannon sampling theorem\cite{eleven}, high frequency data of the image is most likely to get lost while downscaling; which makes the choice of algorithm for downscaling of large size images a crucial part of preprocessing images. Moreover, downscaling of input images should not lead to performance loss in classification process due to loss of some information while downscaling. After multiple experiments we decided to downscale image by $8x$. There are primarily two reasons to downscale images by $8x$ :

\begin{itemize}
    \item While downscaling images by $2x$ and $4x$ times, size of input images were still large and training DL network on those images would require more resources.
    \item On the other hand, downscaling by $16x$ and $32x$ times leads to a greater loss of information and hence features of original image were not retained for proper classification.
\end{itemize}

Six downscaling algorthims : Nearest Neighbour, Bilinear, Bicubic, Lanczos, Rapid detail image preserving\cite{sixteen} (RDIP) and Learned Image Downscaling\cite{one} (LID) are taken into consideration. RDIP and LID are content adaptive downsampling algorithms which retain the content of original images while downsampling. 
\begin{itemize}
    \item \textbf{Nearest Neighbour} algorithm uses the nearest neighbour of the sampled pixel.
    \item \textbf{Bilinear} algorithm takes a weighted average of the $4$ neighborhood pixels\cite{twelve}.
    \item \textbf {Bicubic} algorithm takes a weighted average of the $4 \times 4$ neighborhood pixels\cite{twelve}.
    \item \textbf{Lanczos} algorithm uses either of $4 \times 4$, $6 \times 6$ or $8 \times 8$ window kernel of sinc function to generate the downscaled images\cite{thirteen}.
    \item \textbf{RDIP} algorithm computes downscaled image from its box filtered form. The algorithm gives more weight to pixels that differ from neighborhood pixels in downscaled image\cite{sixteen}.
    \item \textbf{LID} trains a convolutional neural network for generating downscaled and upscaled images of original size. The resampler network in the architecture is responsible for downscaling images. Pretrained model for $4x$ and $2x$ downscaling is used to downscale images by $8x$\cite{one}.
\end{itemize}

\begin{figure}
\begin{center}
  \includegraphics[width=0.95\linewidth]{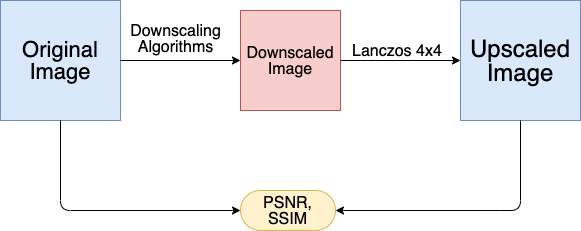}
  \caption{Flowchart for comparing the performance of various downscaling algorithms : Original Image is downscaled $8$ times by using different algorithms and upscaled $8$ times by using Lanczos $4 \times 4$ filter.}
\label{fig1}
\end{center}
\end{figure}

In order to compare the results of different downscaling algorithms we use the flowchart shown in Figure \ref{fig1}. After applying different algorithms to downscale the input image by $8x$, we will upscale those images. The original image and upscaled images are then compared using methods for evaluation of image quality assessment. The upscaling method employed is the use of Lanczos $4 \times 4$ sinc function kernel to upscale $8x$ times\cite{thirteen}. As the comparison is to be made for downscaling algorithms we keep the upscaling algorithm same for all the methods of downscaling.  


To compare the original image ($I_{org}$) and image obtained after downscaling and upscaling ($I_{res}$); we use Peak Signal to Noise Ratio (PSNR) and Structural Similarity Index (SSIM). PSNR value measures how much an image is reconstructed after distortion. SSIM value helps to find how similar the two images are.

\begin{itemize}
    \item \textbf{PSNR} measures image quality difference based on pixel difference. The mathematical formula for PSNR between two images is calculated by Eq. \ref{eq1}\cite{fourteen}.
\begin{itemize}
    \item $x$ and $y$ are input images.
    \item $s$ is $255$ as used for 8-bit input image.
    \item $MSE$ is the mean squared error between images.
\end{itemize}
\begin{equation}
    \label{eq1}
    PSNR(x,y)  = 10 log \frac{s^2}{MSE(x,y)}
\end{equation}
As PSNR is inversely proportional to MSE, more the PSNR means less the MSE and hence better method of downscaling.
    \item \textbf{SSIM} is calculated by applying sliding window of size $8 \times 8$ that moves pixel by pixel to cover all rows and columns of image. SSIM is calculated by using Eq. \ref{eq2}\cite{fourteen}. 
\begin{itemize}
    \item $x$ and $y$ are input images.
    \item $\mu_x$ and $\mu_y$ are means of vector formed $8 \times 8$ pixel images.
    \item $\sigma_{x}$, $\sigma_{y}$ and $\sigma_{xy}$ are co-variances of $x$, $y$ and $x$ and $y$ respectively.
    \item $c_1=(k_1L)^2$ and $c_2=(k_2L)^2$ where $L=7$ for $8$ bit image and $k_1 = 0.01$ and $k_2 = 0.03$.
\end{itemize}
\begin{equation}
\label{eq2}
SSIM(x,y) = \frac{(2 \mu_x \mu_y + c_1)(2 \sigma_{xy} + c_2)}{(\mu_x^2 + \mu_y^2 + c_{1})(\sigma_x^2 + \sigma_y^2 + c_{2})}
\end{equation}
Higher the SSIM value, more is the similarity and hence better is the downscaling algorithm.
\end{itemize}


The average PSNR and SSIM values of $I_{orig}$ and $I_{res}$ method wise are reported in Table \ref{tab1} and Table \ref{tab2} respectively. Table highlights that the highest value for PSNR and SSIM is obtained by LID and Bilinear downscaling algorithms.

\begin{table*}
\begin{center}
\begin{tabular}{|c|c|c|c|c|c|c|}
\hline
Class & Nearest Neighbour & RDIP & Lanczos  & Bicubic & Bilinear &  LID\\
\hline
$0$ & $37.769571$ &  $37.949846$ & $38.050363$ & $38.074301$ & $38.084206$  & \textbf{39.015693}\\
\hline
$1$ & $36.937753$ & $37.120090$ & $37.221959$ & $37.247297$ & $37.410356$  & \textbf{37.711350}\\
\hline
$2$ & $38.488647$ & $38.663193$ & $38.767848$ & $38.785894$ & $37.710555$ & \textbf{38.791407}\\
\hline
$3$ & $37.537186$ & $37.708631$ & $37.808070$ & $37.829595$ & \textbf{37.846174} & $37.777964$\\ 
\hline
$4$ & $38.384808$ & $38.542223$ & $38.612746$ & $38.633373$ & $38.645576$  & \textbf{39.221942}\\
\hline
\end{tabular}
\end{center}
\caption{PSNR value comparison for image downscaling by Nearest Neighbour, RDIP, Lanczos $4 \times 4$ filter, Bicubic, Bilinear and LID.}
\label{tab1}
\end{table*}

\begin{table*}
\begin{center}
\begin{tabular}{|c|c|c|c|c|c|c|}
\hline
Class & Nearest Neighbour & RDIP & Lanczos  & Bicubic & Bilinear &  LID\\
\hline
$0$ & $0.916743$ & $0.918756$ & $0.919373$ & $0.919742$ & $0.920029$ & \textbf{0.933177}\\
\hline
$1$ & $0.904271$ & $0.906823$ & $0.907258$ & $0.907832$ & \textbf{0.908378} & $0.905576$\\
\hline
$2$ & $0.927558$ & $0.929262$ & $0.929633$ & $0.929976$ & \textbf{0.930178} & $0.910066$\\
\hline
$3$ & $0.908891$ & $0.910721$ & $0.911462$ & $0.911916$ & \textbf{0.912355} & $0.910956$\\
\hline
$4$ & $0.935021$ & $0.936309$ & $0.936583$ & $0.936910$ & \textbf{0.937149} & $0.930768$\\
\hline
\end{tabular}
\end{center}
\caption{SSIM value comparison for image downscaling by Nearest Neighbour, RDIP, Lanczos $4 \times 4$ filter, Bicubic, Bilinear and LID.}
\label{tab2}
\end{table*}

Hence, we continue our experiments with images downscaled by LID And Bilinear algorithms and provide a comparison between classification results obtained on images downscaled by both algorithms.

\subsubsection{Padding}
After downscaling, the images are still of different sizes. In order to bring images to a fixed dimension; we choose padding over resizing because of following main reasons :
\begin{itemize}
    \item Padding helps to preserve the features of original image.
    \item Resizing to a fixed size leads to changes in aspect ratio however padding maintains aspect ratio of image. 
\end{itemize}

Hence, we pad all the images with zeros symmetrically to change the dimension of image from its original dimension to $600 \times 600 \times 3$. The primary reasons to select $600 \times 600 \times 3$ image sizes are :
\begin{itemize}
    \item After downscaling $8x$ times, the image sizes varied between $550$ and $600$.
    \item $600 \times 600 \times 3$ images can be divided equally and provided as an input to DL networks (as generally the DL networks take images of sizes $224 \times 224 \times 3$ or $300 \times 300 \times 3$).
\end{itemize}

Hence, to equalize the dimension, images are padded and resized to $600 \times 600 \times 3$. We calculate the dimension left for equating height and width of the image to $600$ each and then add an equal amount of zeros pixel values on either side along height and width. 

After preprocessing we transform images from varying and large sizes to a fixed dimension and a comparatively smaller size of $600 \times 600 \times 3$.

\subsection{DL Architecture}

\begin{figure}
\begin{center}
\includegraphics[width=0.97\linewidth]{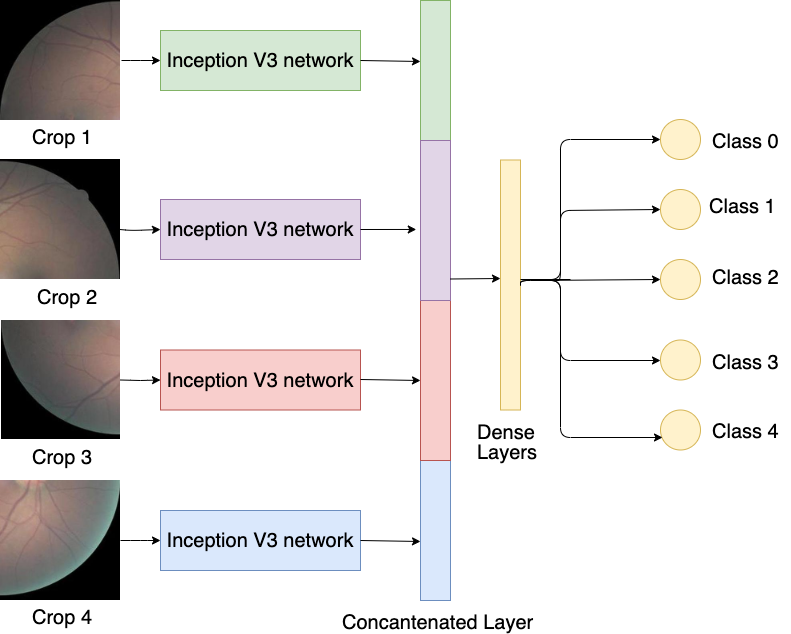}
  \caption{Multi Channel Inception V3 architecture. Four crop portions of $600 \times 600 \times 3$ image fed into Inception V3 which concatenates the results and trains fully connected layers to give $5$ class classification.}
  \label{fig2}
  \end{center}
\end{figure}

We use DL approach for multi class classification. Inception V3 network is trained to classify images. In order to feed input to multi channel network, we divided the preprocessed images into four parts : top left, top right, bottom left and bottom right. Reason behind cropping the image to four part was that our input images are of size $600 \times 600 \times 3$ and Inception V3 network requires input images of size $300 \times 300 \times 3$. So we crop and feed as an input to Multi Channel network as shown in Fig \ref{fig2}. Here Multi Channel Inception V3 architecture is used to address following issues :
\begin{itemize}
    \item The preprocessed images cannot be directly fed as an input to CNN as it increases the number of parameters and computations for a CNN.
    \item In order to use all the features of image for classification, the image is divided into $4$ parts and given to different channel of network to train those features.
\end{itemize}

Transfer learning based approach is used to train the network. In order to check the performance of proposed architecture, accuracy, sensitivity and specificity are used. A brief description about Inception V3, Multi Channel CNN and Transfer Learning is provided in subsequent sections.



\subsubsection{Inception V3}
Inception V3 is a $48$ layer deep convolutional neural network architecture. When applied on ImageNet dataset\cite{nine} to classify images into nearly $1000$ different classes, it shows Top-1 accuracy of $78.8\%$ and Top-5 accuracy of $94.4\%$\cite{eight}. The original $48$ layer network has $23$ million parameters. Here, we remove the fully connected layers and use $42$ layers whose weights are initialized with weights of benchmark ImageNet dataset\cite{nine} results. Inception V3 uses $3 \times 3$, $5 \times 5$ and $7 \times 7$ convolutional kernel filters to extract features from input images. In order to obtain the maximum features from fundus images; we chose to use pretrained Inception V3 network so that weights are not randomly initialized.

\subsubsection{Multi Channel Architecture}
Multi Channel CNN are used to train multiple networks in parallel. The features from the multiple input images are extracted simultaneously in the networks. In the proposed architecture, we provide the cropped images of size $300 \times 300 \times 3$ as an input to the four channels of the architecture. Each channel trains a Inception V3\cite{eight} network. We concatenate the last layers of all Inception V3 network. We introduce fully connected layers after concatenation whose weights are trained to obtain better performance in classification. Figure \ref{fig2} shows the architecture for Class $2$ image of Kaggle dataset\cite{six}. 

\subsubsection{Transfer Learning} 
We use the approach of transfer learning for training the architecture. Transfer learning refers to a method where model trained for a problem is reused as starting point of model to solve different problem.  Here we are retaining the pretrained parameters of the network and training the randomly initialized parameters. So the fully connected layers of Multi Channel Architecture are only trained to obtain better results.



\subsubsection{Evaluation Metric}
\label{sec4}
DR is a multi class classification problem. Hence, to evaluate the performance of our algorithm we report three metrics : accuracy, sensitivity and specificity. While evaluating we consider classes $1$, $2$, $3$ and $4$ as class that have images having DR and class $0$ as a class not having DR. On the basis of this, we construct a confusion matrix as shown in Table \ref{tab7} to calculate the evaluation metrics. In order to compare our results with state of the art methods, the values of the three metrics are required. The method and equations used to calculate them are mentioned below.
\begin{itemize}
    \item \textbf{Accuracy} : Accuracy helps to measure what ratio of fundus images are correctly classified. Accuracy is calculated using Equation \ref{eq5}.
    \begin{equation}
        \label{eq5}
        Accuracy = \frac{TP + TN}{TP + FP + TN + FN}
    \end{equation}
    A higher value of accuracy means having correct classification of data samples\cite{twenty-five}.
    \item \textbf{Sensitivity} : Sensitivity measures what ratio of fundus images classified as having DR actually have DR. Sensitivity is calculated using Equation \ref{eq3}.
    \begin{equation}
        \label{eq3}
        Sensitivity = \frac{TP}{TP + FN}
    \end{equation}
    A higher value of sensitivity means model performed well in detecting the disease in patients correctly\cite{twenty-five}.
    \item \textbf{Specificity} : Specificity measures what ratio of fundus images classified as not having DR actually did not have DR. Specificity is calculated using Equation \ref{eq4}.
    \begin{equation}
        \label{eq4}
        Specificity = \frac{TN}{TN + FP}
    \end{equation}
    A higher value of specificity means model performed well in detecting that the disease was not caused\cite{twenty-five}.
\end{itemize}

\begin{table}
    \begin{center}
    \begin{tabular}{|c|c|c|}
        \hline
         Actual / Predicted & Images not having DR & Images having DR\\
         \hline
         Images not having DR & $TN$ & $FN$\\
         \hline
         Images having DR & $FP$ & $TP$\\
         \hline
    \end{tabular}
    \end{center}
    \caption{Confusion matrix to be referred to calculate accuracy, specificity and sensitivity}
    \label{tab7}
\end{table}
We perform experiments and report accuracy, sensitivity and specificity which outperforms existing methods. 


\section{Experiment Setup}
We train multi channel architecture with amalgamated dataset. The dataset is divided into three parts : train set, test set and validation set. Test set consists of $20\%$ images from all classes. We divide the rest $80\%$ of data into two parts : $80\%$ for training and $20\%$ for validating. Due to high imbalance in the dataset, we divide data into three sets class-wise so that train, test and validation sets do not lack samples from any class. On the basis of results from Table \ref{tab1} and Table \ref{tab2}; we trained two  Multi Channel Inception V3 architecture on downscaled images obtained from Bilinear and LID algorithm.

\section{Results}

As mentioned in Section \ref{sec4}, evaluation of the proposed architecture is done by calculating accuracy, specificity and sensitivity. Table \ref{tab8} and Table \ref{tab9} show confusion matrix obtained by testing two Multi Class Inception V3 networks on images downscaled by Bilinear algorithm and LID respectively. On the basis of results obtained from confusion matrix; we calculate specificity and sensitivity for both methods and report in Table \ref{tab6}. Multi Channel Inception V3 with Bilinear downscaling results in $83.15\%$ accuracy, $81.2\%$ sensitivity and $84.6\%$ specificity and with LID downscaling results in $85.2\%$ accuracy, $83.4\%$ sensitivity and $87.6\%$ specificity. We observe that accuracy, sensitivity and specificity values obtained using LID downscaling algorithm are better compared to that obtained when Bilinear downscaling algorithm is used. 

\begin{table}
    \begin{center}
    \begin{tabular}{|c|c|c|}
        \hline
         Actual / Predicted & Images not having DR & Images having DR \\
         \hline
         Images not having DR & $3112$ & $718$ \\
         \hline
         Images having DR & $564$ & $3216$ \\
         \hline
    \end{tabular}
    \end{center}
    \caption{Confusion matrix for Multi Channel Architecture with Bilinear downscaling}
    \label{tab8}
\end{table}

\begin{table}
    \begin{center}
    \begin{tabular}{|c|c|c|}
        \hline
         Actual / Predicted & Images not having DR & Images having DR \\
         \hline
         Images not having DR & $3194$ & $664$ \\
         \hline
         Images having DR & $454$ & $3298$ \\
         \hline
    \end{tabular}
    \end{center}
    \caption{Confusion matrix for Multi Channel Architecture with LID }
    \label{tab9}
\end{table}

\begin{table}
\begin{center}
\begin{tabular}{|p{3.5cm}|c|c|c|}
    \hline
        Method & Accuracy & Sensitivity & Specificity \\
        \hline
        CNN with $13$ layers\cite{seven} & $75\%$ & $30\%$ & \textbf{95\%} \\
         \hline
         GoogleNet\cite{twenty-eight} & $81$\% & $35$\% & $88$\%\\
         \hline
        Shallow CNN \cite{twenty-six} & $42$\% & NR & NR \\
        \hline
        VGG16 \cite{twenty-six} & $78.3$\% & NR & NR \\
        \hline
        BT-VGG\cite{twenty-three} & $83.2\%$ & $81.8\%$ & $89.3\%$ \\
         \hline
        Multi Channel Architecture with Bilinear downsampling & $83.15\%$ & $81.2\%$ & $84.6\%$ \\
         \hline
         Multi Channel Architecture with LID & \textbf{85.2\%} & \textbf{83.4\%} & $87.6\%$ \\
         \hline
    \end{tabular}
    \end{center}
    \caption{Comparison of results obtained by proposed architecture with state of art methods \\
    (NR = Not reported)}
\label{tab6}
\end{table}

While comparing the results with state of the art methods in Table \ref{tab6}, we observe that Multi Channel architecture with LID algorithm outperforms existing methods by showing $85.2\%$ accuracy and $83.4\%$ sensitivity. However, the specificity obtained from CNN with $13$ layers \cite{seven} is highest compared to all methods. Here, higher value of specificity is not obtained as the network learned on a highly skewed dataset as mentioned in Sec. \ref{sec2}. However, the architecture provided results that beat the benchmark results to detect DR in eyes correctly. Hence, we conclude that the proposed Multi Channel Inception V3 architecture with LID downscaling outperforms the state of art methods.


\section{Conclusion}
Large image sizes are the major issues while dealing with bio-medical image classification problems. Downscaling of images thus becomes an important part of preprocessing before feeding them into training models. On the basis of experiments conducted to calculate the PSNR and SSIM values of reconstructed images; we conclude that amongst all methods Bilinear algorithm and LID gives better results.

On the basis of experiments conducted on images downscaled by Bilinear and LID by using proposed Multi Channel Inception V3 network; we conclude that the results obtained by using LID downscaling algorithm outperforms the results found using Bilinear algorithm in preprocessing. Moreover, the results of experiments also show improvements over existing approaches.

Thus, we propose a novel approach to deal with large and varying size input data by downscaling methods and multichannel CNN. Based on input dataset, modifying downscaling parameters and cropping of images for multichannel CNN one can use this architecture for classification tasks.

\bibliographystyle{IEEEtran}
\bibliography{ref}




\end{document}